\documentclass[A4paper, 11 pt, conference]{ieeeconf}  
\IEEEoverridecommandlockouts       
\usepackage{geometry}
\usepackage{graphics} 
\usepackage{epsfig} 
\usepackage{mathptmx} 
\usepackage{times} 
\usepackage{amsmath} 
\usepackage{amssymb}  
\usepackage{subfigure}
\usepackage{algorithm} 
\usepackage{algorithmic} 
\usepackage{multirow} 
\begin{document}
	
	\title{\LARGE \bf
		Contrastive Disentangled Learning on Graph \\ for Node Classification}
	
	\author{Xiaojuan Zhang, Jun Fu$^*$, \IEEEmembership{Senior Member, IEEE,} Shuang Li 
		\thanks{Xiaojuan Zhang is with the State Key Laboratory of Synthetical Automation for Process Industries, Northeastern University, Shenyang, China. (e-mail: XiaojuanZhang123@gmail.com).}
		\thanks{Jun Fu is the corresponding author and with the State Key Laboratory of Synthetical Automation for Process Industries, Northeastern University, Shenyang, China. (e-mail: junfu@mail.neu.edu.cn)}
		\thanks{Shuang Li is with School of Computer Science and Technology, Beijing Institute of Technology, Beijing, China. (e-mail:  shuangli@bit.edu.cn)}
	}
	
	\maketitle 
	\thispagestyle{empty}
	\begin{abstract}
		Contrastive learning methods have attracted considerable attention due to their remarkable success in analyzing graph-structured data.  Inspired by the success of contrastive learning, we propose a novel framework for contrastive disentangled learning on graphs, employing a disentangled graph encoder and two carefully crafted self-supervision signals. Specifically, we introduce a disentangled graph encoder to enforce the framework to distinguish various latent factors corresponding to underlying semantic information and learn the disentangled node embeddings. Moreover, to overcome the heavy reliance on labels, we design two self-supervision signals, namely node specificity and channel independence,  which capture informative knowledge without the need for labeled data, thereby guiding the automatic disentanglement of nodes. Finally, we perform node classification tasks on three citation networks by using the disentangled node embeddings, and the relevant analysis is provided. Experimental results validate the effectiveness of the proposed framework compared with various baselines.
		
	\end{abstract}
	\section{Introduction} \label{Introduction}
	In recent years, with the explosive growth of graph data in various fields such as social networks \cite{daud2020applications}, recommendation systems \cite{DBLP:conf/cikm/WuSCCLZW021}, and bioinformatics \cite{Lv2021}, the need for effective graph representation learning has become more urgent. Graph representation learning methods aim to learn the low-dimensional vector representations that capture structural and semantic information in graphs. Specifically,  graph convolutional networks (GNNs) \cite{kipf2017semisupervised,Velickovic2018,10.5555/3294771.3294869} have achieved remarkable success in graph representation learning.
	
	However, existing GNNs heavily rely on manually annotated labels, which can be expensive, time-consuming, or even unattainable for many real-world graph datasets. To address this challenge, contrastive learning \cite{You2020,zhu2021graph,Suresh2021} has emerged as a promising unsupervised learning approach for graph representation learning. Contrastive learning leverages the intrinsic structure of the data to learn meaningful representations without the need for explicit labels. Nevertheless, the existing methods usually adopt a holistic strategy that treats the node's neighborhood as a perceptual whole. This strategy overlooks the subtle differences among different parts of the neighborhood, which are crucial for capturing the underlying structure and semantics of the graph. 
	
	It is worth noting that real-world graphs are often formed through complex and heterogeneous processes influenced by multiple latent factors \cite{Ma2019, Liu2020, zhao2022exploring}. For example, a person's social network may have several different components. To be specific, a person may have a group of friends from their hometown, colleagues from their current job, and classmates from a course they are taking. Each of these groups connects with the person for different reasons, so node representations in the network may be disentangled into several components corresponding to different latent factors.
	
	Motivated by the above analyses, in this paper, we propose a new Contrastive Disentangled Learning framework on Graph (CDLG). The framework incorporates a disentangled graph encoder and two self-supervision signals. By leveraging a neighborhood routing mechanism, the disentangled graph encoder can identify diverse latent factors, leading to disentangled node embeddings. Then, two self-supervision signals, namely node specificity and channel independence, are designed to guide the model training and optimization process. Subsequently, the learned disentangled embeddings are fed into a linear classifier for downstream tasks, specifically node classification tasks. Comparative experiments against several baseline models demonstrate that the proposed approach achieves substantial performance improvements on three citation networks.
	
	Our main contributions of this paper can be summarized as follows:
	\begin{itemize}
		\item {\textbf{Contrastive Disentangled Learning Framework on Graphs} 
			\\ The paper proposes a novel framework called CDLG, which combines a disentangled graph encoder and two self-supervision signals. This framework addresses the challenge of entangled embeddings in graph data and enables the learning of disentangled node embeddings.}
		\item {\textbf{Self-supervision signals without labeling dependency.} 
			\\ Our carefully designed self-supervision signals node specificity and channel independence, allow our method to overcome the dependence on manual labels, thereby reducing costs. }
		\item {\textbf{Performance Improvement} 
			\\ Through extensive experiments on three citation networks, the paper demonstrates the superiority of the CDLG framework compared to several baseline models in node classification tasks.}
	\end{itemize}
	This paper is structured as follows:
	Section \ref{Introduction} provides a brief introduction, outlining the motivation and overview of the proposed approach.
	Section \ref{Related Work} discusses related work in the field.
	In Section \ref{CLDGE: the Proposed Model}, we present the CDLG method in detail.
	Node classification experiments are conducted in Section \ref{Experiments} to showcase the superior performance of our model.
	Finally, Section \ref{Conclusions} concludes the paper, summarizing the key findings and discussing potential future research directions.
	\section{Related Work} \label{Related Work}
	\subsection{Graph Contrastive Learning}
	Recently, contrastive learning methods \cite{Velickovic2019,qiu2020gcc,Zhu2020,DBLP:conf/iclr/SunHV020} which incorporate various designs of pretext tasks have become mainstream methods in graph self-supervised learning domain. These methods aim to learn a similarity metric between pairs of views generated by graph data augmentation approaches. Existing methods such as DeepWalk \cite{DBLP:conf/kdd/PerozziAS14} and node2vec \cite{grover2016node2vec} employ skip-gram models to learn node representations through truncated random walks. GRACE \cite{Zhu2020} contrasts two views between different nodes to construct pretext tasks. MVGRL \cite{hassani2020contrastive} and GCA \cite{zhu2021graph} contrast different views between nodes and subgraphs. DGI \cite{Velickovic2019} contrasts views between nodes and the entire graph.
	However, the existing methods suffer from the problems of undiscovered underlying information and severe entanglement among latent factors. Consequently, they may fail to identify and infer various latent factors behind complex graph data and easily lead to suboptimal embeddings for downstream tasks. Therefore, there is an urgent need for new methods that can effectively disentangle the latent factors within graph data.
	
	\subsection{Graph Disentangled Representation Learning}
	Graph disentangled representation learning aims to decompose the feature representations of graph data into several components that correspond to specific underlying latent factors. 
	Ref.\cite{Ma2019} is the first work that proposes disentangled graph convolutional networks (DisenGCN). DisenGCN employs the neighborhood routing mechanism to achieve node disentanglement between different latent factors. 
	To address this issue, subsequent works \cite{Liu2020,Zheng2021} have expanded upon DisenGCN for node disentanglement. IPGDN \cite{Liu2020} incorporates the Hilbert Schmidt Independence Criterion (HSIC) into DisenGCN to promote independence between the latent representations. ADGCN \cite{Zheng2021} introduces an adversarial regularizer to enhance the separability among the component distributions. Furthermore, recent studies have explored disentanglement in terms of subgraph \cite{yang2020factorizable}, edge \cite{zhao2022exploring}, and multi-level information \cite{wu2022multi}. Nevertheless, most existing methods heavily rely on annotated labels, which can be expensive, time-consuming, or even unattainable, especially when dealing with large-scale datasets \cite{Liu2022}. To overcome these challenges, this paper proposes a novel contrastive learning framework that extracts information through well-designed self-supervision signals, leading to improved node embedding representations.
	\begin{figure}[tb]
		\centering
		\includegraphics[width=1\linewidth]{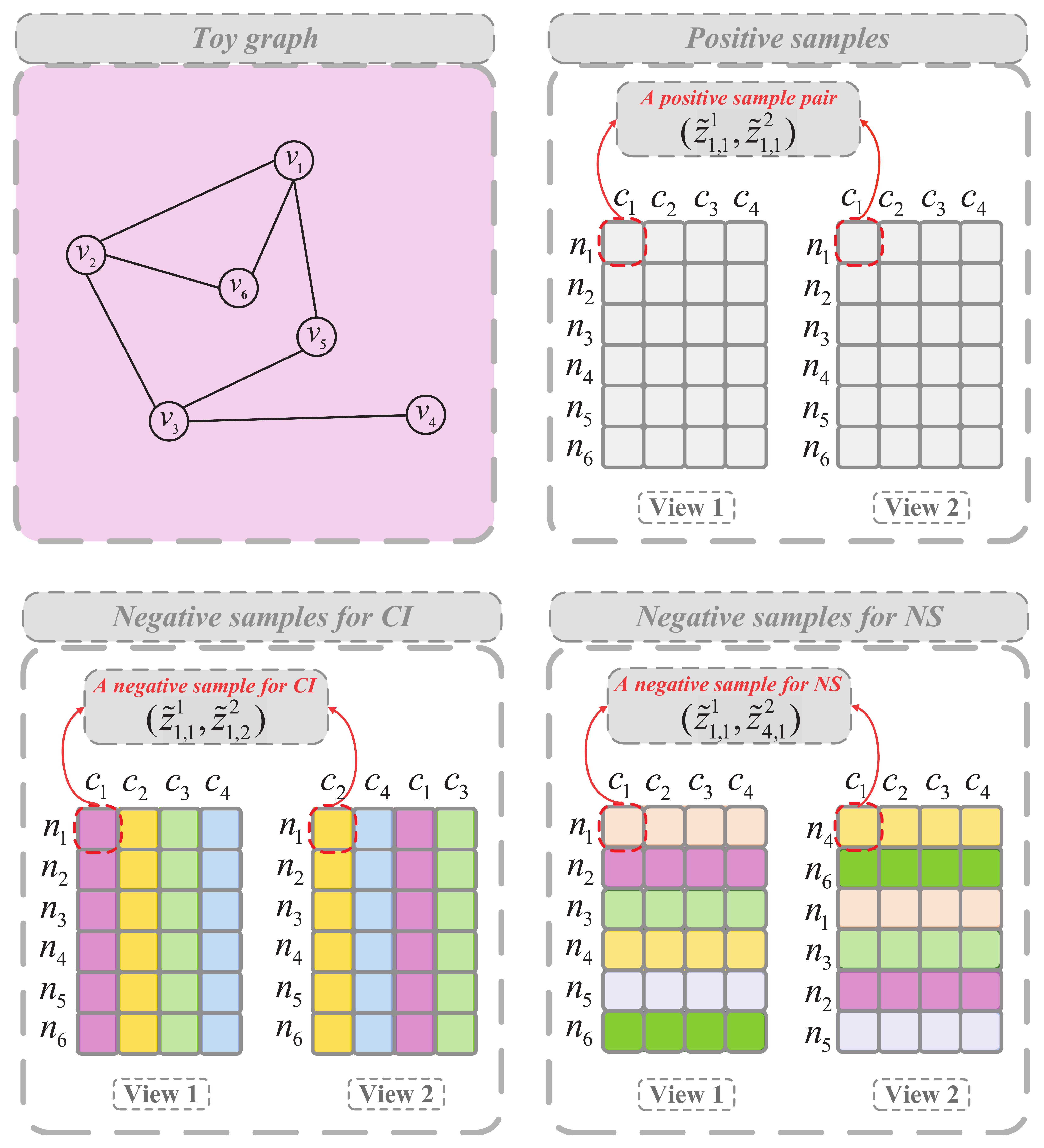}
		\caption{A toy graph to illustrate sampling pairs.}	
		\label{fig:toy graph}
	\end{figure}
	\section{CDLG: the Proposed Model} \label{CLDGE: the Proposed Model}
	\subsection{Preliminaries}
	Let $G=\left \{ V, E, X\right \}$ be an attributed graph, where $V=\left\{ v_{1}, v_{2}, \cdots , v_{n}\right\}$ denotes the sets of nodes. $E \subseteq V \times V$ is the set of edges, $X \in \mathbb{R}^{n \times f}$ is the feature matrix which describes the attribute information. An adjacency matrix $A \in \mathbb{R}^{n \times n}$ depicts graph topology, where $A_{i,j}=1$ if there exists an edge between $v_{i}$ and $v_{j}$, otherwise $A_{i,j}=0$.
	
	In this paper, our goal is to learn generalized and disentangled node representations to benefit the node classification task  \cite{kipf2017semisupervised}. $Y \in \mathbb{R}^{n}$ is the class information for the nodes in $G$. In addition, we assume that the node representations learned by disentangled graph encoder consist of $K$ components corresponding to $K$ latent factors. In detail, for a given node $u$ with the feature vector $x_{u} \in \mathbb{R}^{f}$, the embedding vector is $z_{u}=\left[ z_{u,1}, z_{u,2}, \cdots, z_{u,K} \right]$, where $z_{u} \in \mathbb{R}^{K \times d}$, $d \leq f$. $z_{u,k}\in \mathbb{R}^{d}$ denotes the embedding related to $k$-th latent factors, namely $k$-th channel.
	\subsection{Disentangled Graph Encoder}
	To achieve disentangling, we project the original node representations into different subspaces, i.e., different channels, which can capture mutually exclusive semantic information. We denote the neighbors of node $u$ as: $N(u)=\left \{v|(u,v) \in G  \right \}$. The initial embedding vector for $k$-th channel can be obtained by the following equations:

	\begin{equation}
		c_{u,k} = \sigma(W_{k}^{\top }x_{u}+b_{k}),
		\label{eq1}
	\end{equation}
	where $W_{k} \in \mathbb{R}^{F \times d}$ and $b_{k} \in \mathbb{R}^{d}$ are the parameters of channel $k$, $\sigma \left ( \cdot  \right )$ is a nonlinear activation function, such as $sigmoid(\cdot)$, $relu(\cdot)$. 
	
	The initial embedding vector only utilizes the attribute information of the graph. Furthermore, to take full advantage of the topology of the graph, the neighborhood routing mechanism \cite{Ma2019} is adopted to aggregate the neighborhood information of nodes for each channel:
	\begin{equation}
		z_{u,k}^{(t)} = c_{u,k}+\sum_{v:(u,v) \in G}p_{v,k}^{(t-1)}c_{v,k}
		\label{eq2}
	\end{equation}
	\begin{equation}
		p_{v,k}^{(t)} =\frac{exp(c_{v,k}^{\top}z_{u,k}^{(t)})}{\sum_{k=1}^{K}exp(c_{v,k}^{\top}z_{u,k}^{(t)})} 
		\label{eq3}
	\end{equation}
	where $p_{v,k}$ is the probability that neighbor node $v$ is assigned to the $k$-th channel. $t=1,2,\cdots,T$ is the iterations for the model to converge. To ensure numerical stability, we perform the $l_{2}$-normalization for $z_{u,k}$ before it is used to calculate probability $p_{v,k}$.
	
	\subsection{The Contrastive Learning Framework} After computing the embedding representations $z_{u,k}$ of each channel, we concatenate all $K$ channels to obtain the final embedding vector for node $u$, i.e. $z_{u}=\left[ z_{u,1}, z_{u,2}, \cdots, z_{u,K} \right]$.
	We first generate two graph views that are closely related to the quality of the learned embeddings for our contrastive learning method.
	We adopt two well-known and widely used data augmentation approaches, i.e., removing edges (RE) \cite{Zhu2020} and masking node features (MF) \cite{Hu2020}. It is worth noting that the approach involves the attribute level information and the topological level simultaneously.
	
	Then, we design two self-supervision signals to construct the contrastive loss function. 
	Fig.\ref{fig:toy graph} shows how these two signals guide sampling. 
	The toy graph has 6 nodes $V=\left \{ v_{1}, \cdots, v_{6}\right \}$. The embedding matrices of the two views are generated respectively by the disentangled graph encoder. For the sketch map of each view, each row corresponds to a node, denoted as $n_{i}$ with $i=1,\cdots,6$ , and each column corresponds to a channel, denoted as $c_{j}$ with $j=1,2,3,4$ for the toy graph. Each square block of the view represents an embedding vector, denoted as $\tilde{z}_{i,j}^{m}$, which indicates the $i$-th node embedding vector of $j$-th channel in $m$-th view.
	
	Let input a graph $G=\left (X, A \right )$ with node set $V$ and $|V|=n$. We generate two graph views $\tilde{G}_{1}$ and $\tilde{G}_{2}$ through the generator $\tilde{G}_{i}=\left (\varphi (X),  \psi (A)\right )$. For any node $u\in V$, the first view embedding set in $k$-th channel for graph $S_{1,k}=\tilde{G}_{1}$ is $\left \{\tilde{z}_{u,k}^{1}|u\in V \right \} $. The second view embedding set in $k$-th channel for graph $\tilde{G}_{2}$ is $S_{2,k}=\left \{\tilde{z}_{u,k}^{2}|u\in V \right \} $. We denote the positive sample set for $k$-th channel as $S_{p}=\left \{ \left (\tilde{z}_{u,k}^{1},  \tilde{z}_{u,k}^{2} \right )  \right \}_{n}$ with $k=1,2,\cdots, K$. 
	
	We use noise-contrastive type objectives with standard binary cross-entropy (BCE) loss to maximize the probability of positive samples and minimize the probability of negative samples. The loss function for the positive samples is:
	\begin{equation}
		\mathcal{L}_{p}=-\frac{1}{n \times K}\left [ 
		\sum _{k=1}^{K}\sum _{u\in V}log\left ( \mathcal{D}\left ( \tilde{z}_{u,k}^{1}, \tilde{z}_{u,k}^{2} \right ) \right )
		\right ],
		\label{eq4}
	\end{equation}
	where $\mathcal{D}\left ( \cdot \right )$ is a discriminator that estimates the agreement score between two views. Here, we implement the evaluation by using cosine similarity.
	
	Negative sample pairs are provided by the following two self-supervision signals:
	\subsubsection{Node Specificity (NS)}
	We devise the first signal to enforce that for the same node, the learned embeddings of two views are consistent with each other. For different nodes, the learned embeddings are able to capture discriminative information and be distinguished from each other. To achieve that, a negative sampling strategy is implemented as follows. The order in $S_{1,k}=\left \{\tilde{z}_{u,k}^{1}|u\in V \right \} $ is treated as the anchor, the order in $S_{2,k}=\left \{\tilde{z}_{u,k}^{2}|u\in V \right \} $ is shuffled. Moreover, we denote the NS negative sample set as $S_{ns}=\left \{  \left (\tilde{z}_{u,k}^{1}, \tilde{z}_{v,k}^{2}   \right )   \right \}_{n} $, where $u\neq v$ and $u,v \in V$. The loss function for the NS negative sample pairs is:
	\begin{equation}
		\mathcal{L}_{ns}=\!-\! \frac{1}{n \times K}\left [ 
		\sum _{k=1}^{K}\sum _{S_{ns}} \left (  1-log\left ( \mathcal{D}\left ( \tilde{z}_{u,k}^{1}, \tilde{z}_{v,k}^{2} \right ) \right )\right )
		\right ]
		\label{eq5}
	\end{equation}
	\begin{table}[t!]
		\caption{Dataset statistics}
		\centering
		\begin{tabular}{lccc}
			\hline\hline
			Datasets  &Nodes   &Links   &Features \\
			\hline
			Cora      &2708    &5278    &1433      \\
			Citeseer  &3327    &4552    &3703      \\
			Pubmed    &19717   &44324   &500       \\      	
			\hline
		\end{tabular}	
		\label{tab:Dataset statistics}
	\end{table}
	\subsubsection{Channel Independence (CI)}
	The signal is designed to restrict the independence between various channels because distinct channels are supposed to capture exclusive and non-overlapping information. In this case, for any node $u$, we treat the order of $S_{1,k}$ as the anchor, and shuffle the order of channels for the second view. The CI negative sample set is denoted as $S_{ci}=\left \{  \left (\tilde{z}_{u,k}^{1}, \tilde{z}_{u,m}^{2}   \right )   \right \}_{n \times K} $, where $k\neq m$ and $k,m =1,2,\cdots, K$. The loss function for the negative sample pairs of CI is:
	\begin{equation}
		\mathcal{L}_{ci}=-\frac{1}{n \times K}\left [ 
		\sum _{S_{ci}} \left (  1-log\left ( \mathcal{D}\left ( \tilde{z}_{u,k}^{1}, \tilde{z}_{u,m}^{2} \right ) \right )\right )
		\right ]
		\label{eq6}
	\end{equation}
	\subsection{Optimization}
	In summary, the overall objective function for optimization  is defined as follows:
	\begin{equation}
		\mathcal{L}=\mathcal{L}_{p}+\mathcal{L}_{ns}+\mathcal{L}_{ci}
		\label{eq7}
	\end{equation}
	After optimizing the disentangled graph encoder based on the contrastive learning framework to extract features, a $l_{2}$-regularized logistic regression classifier is trained to perform node classification tasks, following the setup in \cite{Velickovic2019}.
	\subsection{Optimization}
	
	\section{Experiments} \label{Experiments}
	In this section, we conduct extensive experiments to evaluate the quality of disentangled node embeddings in the way of performing node classification tasks on three citation network datasets.
	\begin{table}[!t]
		\caption{Results for node classification}
		\centering
		\begin{tabular}{lcccc}
			\hline\hline
			Methods  &Training Data   &Cora   &Citeseer   &Pubmed \\
			\hline	   
			DeepWalk   &X       &67.2 &43.2 &65.3      \\
			GraphSAGE  &X,A     &78.7 &69.4 &78.1      \\
			MVGRL      &X,A     &82.9 &72.6 &79.4      \\     
			GRACE      &X,A     &80.0 &71.7 &79.5      \\  
			DGI        &X,A     &82.3 &71.8 &76.8      \\ 
			\hline
			DisenGCN   &X,A,Y   &83.7 &73.4 &80.5      \\    
			\hline\hline
			CDLG     &X,A     &82.5 &73.6 &81.5      \\   	
			\hline
		\end{tabular}	
		\label{tab:results}
	\end{table}
	\subsection{Experimental Setup}
	We make use of three widely-used citation networks (Cora, Citeseer, Pubmed) \cite{Yang2016}, where nodes correspond to articles and edges relate to citation links between articles. In addition, we implement all experiments by using PyTorch 1.12.1 on NVIDIA GeForce RTX 3090.
	
	Furthermore, we compare CDLG with three types of baselines for comprehensive evaluation. The first type consists of contrastive learning-based methods, including DeepWalk \cite{DBLP:conf/kdd/PerozziAS14}, MVGRL \cite{hassani2020contrastive}, GRACE \cite{Zhu2020}, and DGI \cite{Velickovic2019}. The second type comprises classical supervised graph representation learning methods, such as GraphSAGE \cite{DBLP:conf/cikm/WuSCCLZW021}. Additionally, we also include DisenGCN \cite{Ma2019}, a supervised learning method that employs disentangled node embeddings, for comparison.
	
	For the node classification task, we follow the standard split in \cite{kipf2017semisupervised}, where 20 nodes per class are used for training, and the validation set and the test set contain 500/1000 nodes respectively. The parameters are optimized with Adam \cite{Kingma2015}, where the learning rate for the encoder is 0.001 and for the classifier is 0.01. We use $d=32$ for each channel and tune the hyper-parameters by grid search, where the number of channels $K$ are searched from $\left \{ 2,4, \cdots, 16  \right \}$, the ratio of removing edges, and the ratio of masking node features (MF) are searched from $\left \{ 0.1, 0.2, \cdots , 1\right \}$.
	
	\begin{figure}[t]
		\centering
		\includegraphics[width=1\linewidth]{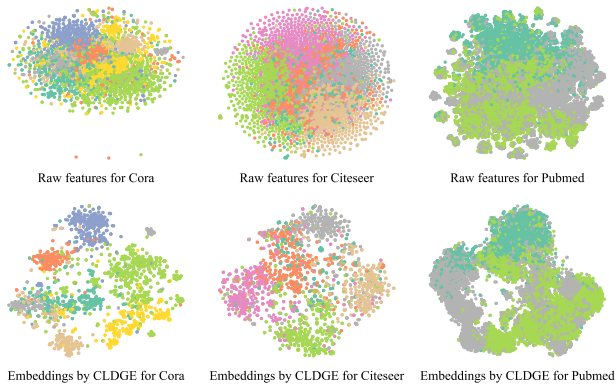}
		\caption{Visualization comparison.	}
		\label{fig:tsne}
	\end{figure}
	\subsection{Results}
	The results for the node classification task are summarized in Table.\ref{tab:results}, where the performance is evaluated in terms of accuracy. It is obvious that our proposed CDLG method consistently outperforms the unsupervised baselines, achieving a margin of 1\% higher accuracy on the Citeseer dataset and 2\% higher accuracy on the Pubmed dataset.  Even though DisenGCN is the supervised method that utilizes the labels $Y$, CDLG still matches its performance on Citeseer and exceeds 1\% on the Pubmed dataset. On the Cora dataset, the performance of CDLG is slightly lower than MVGRL among the unsupervised approaches. Despite a 1.2\% gap in performance between the proposed method and DisenGCN on Cora, the absence of manually annotated labels makes CDLG a more cost-effective option compared to DisenGCN.
	
	\begin{figure}
		\centering
		\subfigure[Loss on Cora]{
			\begin{minipage}[t]{0.45\linewidth}
				\centering
				\includegraphics[width=1.5in]{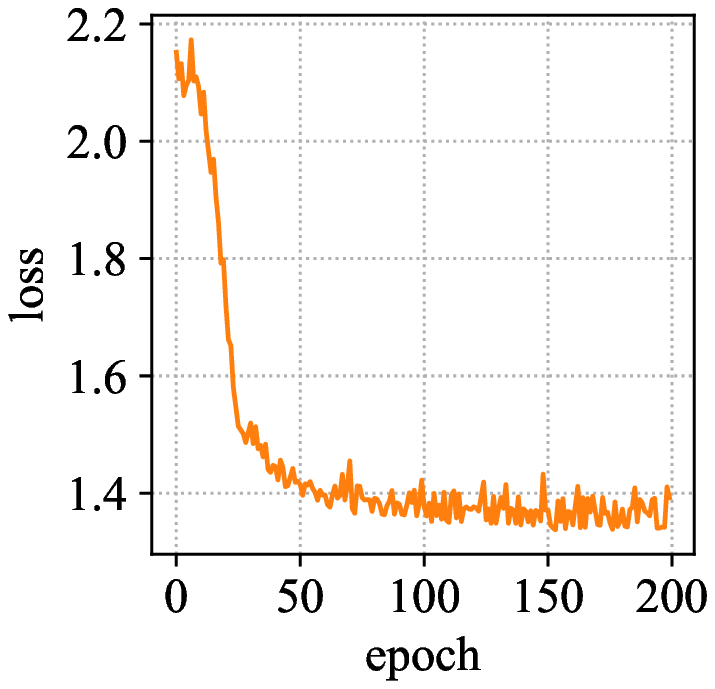}
			\end{minipage}
		}%
		\subfigure[Loss on Citeseer]{
			\begin{minipage}[t]{0.45\linewidth}
				\centering
				\includegraphics[width=1.5in]{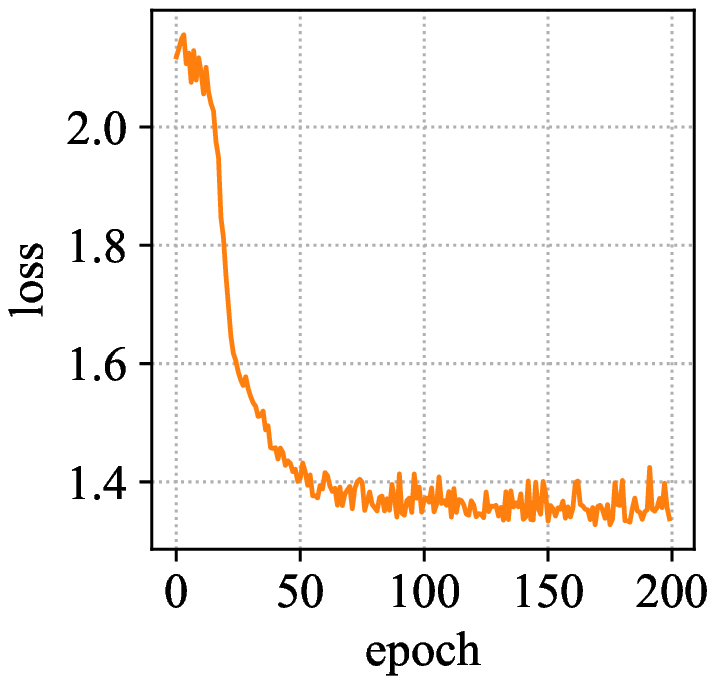}
			\end{minipage}
		}%
		
		\subfigure[Loss on Pubmed]{
			\begin{minipage}[t]{0.45\linewidth}
				\centering
				\includegraphics[width=1.5in]{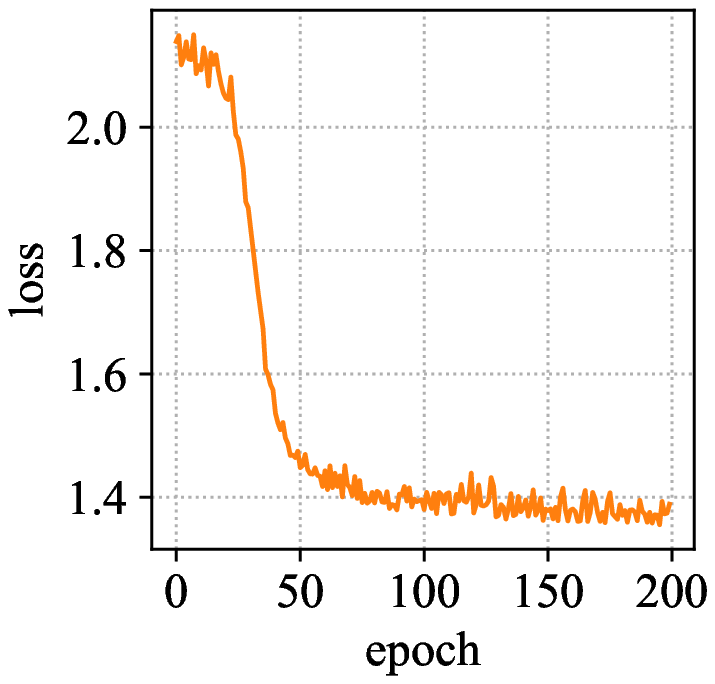}
			\end{minipage}
		}%
		\centering
		\caption{Convergence results on datasets.}
		\label{fig2}
	\end{figure}
	\subsection{Analysis}
	To demonstrate the efficiency of our model, we perform the analysis of the training loss in Fig.\ref{fig2}, which shows that CDLG converges within a small number of iterations.
	
	In addition, to present a more intuitive perspective, we visualize the raw feature representations and the embedding representations of three datasets in a two-dimensional space by applying the $t$-SNE algorithm \cite{VanderMaaten2008}. It is worth noting that a more meaningful layout of the graph data is presented compared to the raw feature representations in Fig.\ref{fig:tsne}.
	\section{Conclusions} \label{Conclusions}
	In this paper, we present a novel contrastive disentangled learning framework for obtaining disentangled node embeddings in graphs. Our model utilizes a disentangled graph encoder with the neighborhood routing mechanism to learn node embeddings. Additionally, we introduce two self-supervision signals to guide the optimization of the model. Through extensive experiments on three citation network datasets, we demonstrate the effectiveness of our approach compared to several baselines for node classification tasks.
	
	In the future, efforts can be done to design a multi-level contrastive disentangled learning framework. For instance, to further enhance our understanding and learning of the intricate underlying factors in graphs, an interesting direction would be to extend the concepts proposed in this paper from nodes to edges or subgraphs. This expansion would enable a more comprehensive analysis of the graph structure and semantics, potentially leading to improved performance in various downstream tasks. 
	\section*{Acknowledgment}
	This work was supported in part by the National Natural Science Foundation of China under Grants 61825301.

\end{document}